\documentclass[runningheads]{llncs}

\usepackage[T1]{fontenc}

\usepackage{graphicx}
\usepackage{color}
\usepackage[marginal]{footmisc}

\usepackage{multicol}
\usepackage{booktabs} 
\usepackage{float}
\usepackage{multirow}
\usepackage{arydshln}
\usepackage[table,xcdraw]{xcolor}
\usepackage{amssymb}
\usepackage{mathrsfs}
\usepackage{bbding}
\usepackage{pdfpages}
\usepackage{tikz}
\usepackage{graphicx}
\usepackage{caption}
\usepackage{bbding}
\usepackage{microtype}
\usepackage{hyperref}

\begin{document}

\begin{sloppypar}

\title{Diverse Generation while Maintaining Semantic Coordination: A Diffusion-Based Data Augmentation Method for Object Detection}

\titlerunning{Data Augmentation with Diverse Generation and Semantic Coordination}


\author{Sen Nie$^{\dag}$
\and
Zhuo Wang$^{\dag}$
\and
Xinxin Wang
\and
Kun He*
}

\authorrunning{S. Nie et al.}

\institute{WuHan University, China}

\maketitle     
\footnote{\dag \ Equal contribution, {*}Corresponding author \\}


\begin{abstract}
Recent studies emphasize the crucial role of data augmentation in enhancing the performance of object detection models. 
However,
existing methodologies often struggle to effectively harmonize dataset diversity with semantic coordination.
To bridge this gap, 
we introduce an innovative augmentation technique leveraging pre-trained conditional diffusion models to mediate this balance. 
Our approach encompasses the development of a Category Affinity Matrix, meticulously designed to enhance dataset diversity, 
and a Surrounding Region Alignment strategy, which ensures the preservation of semantic coordination in the augmented images. 
Extensive experimental evaluations confirm the efficacy of our method in enriching dataset diversity while seamlessly maintaining semantic coordination.
Our method yields substantial average improvements of $+1.4$AP, $+0.9$AP, and $+3.4$AP over existing alternatives on three distinct object detection models, respectively.
\end{abstract}


\section{Introduction}
A substantial training dataset is crucial for fully unleashing the potential of deep neural networks~\cite{li2023semantic}. 
Extensive research underscores the critical role of expansive datasets in bolstering the performance of object detection models~\cite{wang2023object,wu2023aligning}. 
However, the process of collecting and annotating large, diverse datasets is often time-consuming and labor-intensive. 
To address this challenge, 
researchers have explored various data augmentation strategies to enrich and diversify the available data resources.

In object detection, 
data augmentation methods can be broadly classified into two categories~\cite{xu2023comprehensive}: 
$\textbf{\textit{(1)}}$ Model-free methods,
which employ direct geometrical transformations to augment the dataset,
with techniques such as Random Erasing~\cite{zhong2020random}, Mixup~\cite{zhang2017mixup} and Mosaic~\cite{ge2021yolox}.
These methods primarily introduce slight modifications, 
preserving the semantic coordination of the entire image but often at the expense of limited diversity.
$\textbf{\textit{(2)}}$ Model-based methods, which utilize image generation models to synthesize images.
A subset of these Model-based methods is capable of enhancing diversity by generating images that introduce new classes. 
For example, 
X-Paste~\cite{zhao2023x} leverages CLIP~\cite{radford2021learning} to extract masks and enhance data through the random insertion of new objects.
InstaGen~\cite{feng2024instagen} randomly amalgamates base and new classes to produce data.
Although these Model-based methods excel in diversifying the dataset, 
they encounter challenges in maintaining the semantic coordination that is crucial for improving inference capabilities, as demonstrated in previous studies ~\cite{li2022scan,oreski2023yolo,wang2023context,zhu2021semantic}.

\begin{figure}[t]
    \centering
    \includegraphics[width=\textwidth]{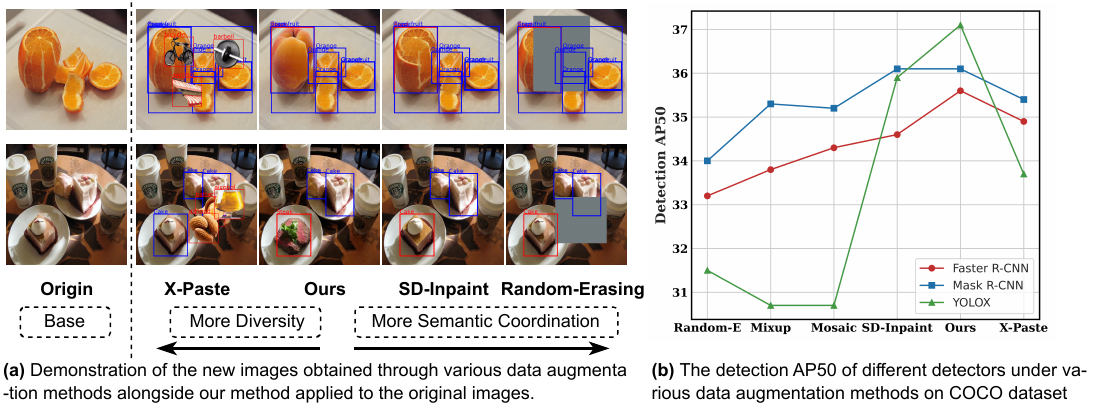}
    \caption{Visual presentation of obtained images and experimental results. Our data augmentation method strikes a balance between semantic coordination and image diversity, leading to the highest performance improvement on AP50.}
    \label{fig:first_page_pic}
\end{figure}

In this paper, we propose a novel method to enhance dataset diversity while maintaining semantic coordination in images. 
Our method leverages a conditional diffusion generative model~\cite{song2020denoising,zhang2023adding} to facilitate modifications of the original images. 
To boost dataset diversity, 
we develop the Category Affinity Matrix, which quantifies visual and semantic similarities across different categories.
The matrix is instrumental in guiding the generative model to produce inter-class images, facilitating the replacement of original objects with those that share an affinity, thereby enriching the diversity of the dataset appropriately.
To address the potential semantic disconnection between objects generated by diffusion models and their background, we introduce the strategy of Surrounding Region Alignment. 
This strategy maintains semantic coordination extracting information from the original diffusion process by DDIM inversion and combining it with information from the new diffusion process, ensuring the semantic integrity of the images produced.
Additionally, inspired by previous work~\cite{suri2023gen2det}, 
we incorporate an Instance-Level Filter to filter out low-quality images.

In our study, We employed six data augmentation methods on three distinct detector architectures: Faster R-CNN~\cite{ren2016faster}, Mask R-CNN~\cite{he2018mask}, and YOLOX~\cite{ge2021yolox}, utilizing three benchmark datasets.
Our method improves the performance of different detector architectures on all datasets and achieves better performance over the existing methods.
Especially, our method leads to $+3.4$AP, $+2.3$AP, and $+6.4$AP gains across different detector architectures compared to the non-augmentation way. Moreover, our method exhibited robust applicability to category-specific and fine-grained datasets, with enhancements of $+3.6$AP and $+4.4$AP, as detailed in Section~\ref{Main Results}. 
These results are further discussed by the detailed analysis provided in Section~\ref{Diversity and Semantic Coordination}, where we delve into the method's ability to strike a balance between diversity and semantic coordination (Fig.~\ref{fig:first_page_pic}).

Our contributions are summarized as follows:
\begin{itemize}
    \item We propose a new diffusion-based data augmentation method to improve the performance of object detection detectors.  
    \item We design the Category Affinity Matrix to generate inter-class images for appropriately enhancing the dataset diversity.
    \item We introduce the Surrounding Region Alignment strategy to maintain semantic coordination across the whole image.
    \item Our methodology achieves notable improvements over existing methods, registering average advantages of $+1.4$AP, $+0.9$AP, and $+3.4$AP across three detector architectures. 
\end{itemize}


\section{Related Work}

\subsection{Data Augmentation for Object Detection}

Data augmentation is to bolster the performance of object detection models by extending datasets. 
Different from simply enlarging the number of samples, bounding box locations and the size of the objects should be considered in the data augmentation for object detection.
Diverse augmentation methods have been explored, ranging from simple transformations such as rotation and flipping to more sophisticated methods including image erasing~\cite{chen2020gridmask,zhong2020random} or image blending~\cite{zhang2017mixup} or image stitching~\cite{yun2019cutmix}. These methods aim to expand datasets with easily calculable bounding boxes while preserving the semantic content of the original image, enhancing model training. However, their diversity is constrained by minor image modifications.
Recent advancements in data augmentation for object detection have transformed into the exploitation of generative models to synthesize realistic training samples. For example, layout-to-image models~\cite{cheng2023layoutdiffuse,inoue2023layoutdm} are trained on datasets with a dense bounding box distribution to generate data. Diffusion models~\cite{ge2022dall,zhao2023x} are utilized to generate images of target objects solely based on textual guidance, and then off-the-shelf segmentation models~\cite{luddecke2022image,qi2022open,qin2020u2} are employed to extract the objects and integrate them into origin images. InstaGen~\cite{feng2024instagen} randomly amalgamates base and new classes and aligns the text embedding of category names with the regional visual feature of the diffusion model to produce data. These methods aim to enhance dataset diversity to improve the robustness and adaptability of model training. However, they do not address semantic coordination issues, which leaves room for improvement.

\subsection{Diffusion Models for Controllable Image Generation.}

Diffusion models have garnered considerable attention due to their capability to generate high-fidelity images while offering explicit control over various aspects of the generation process. In the recent literature, some pivotal and widely utilized models such as GLIDE~\cite{nichol2021glide}, Imagen~\cite{saharia2022photorealistic}, DALL-E 2~\cite{ramesh2022hierarchical}, Latent Diffusion Model (LDM)~\cite{rombach2022high} generate images from text prompts. Meanwhile, further improvements have been made by introducing various conditions to guide more controlled generation and editing. For example, ControlNet~\cite{zhang2023adding} introduces an innovative architecture by incorporating an additional encoder to control diffusion models with spatial signals. IP-Adapter~\cite{ye2023ip} separates cross-attention layers for text features and image features to generate a similar image from multiple perspectives by using an exemplar image as the prompt. ZestGuide~\cite{couairon2023zeroshot} utilizes segmentation maps extracted from cross-attention layers, aligning generation with input masks through gradient-based guidance during denoising. 
In this paper, we also utilize a similar pre-trained state-of-the-art diffusion model for high-fidelity and controllable image generation.

\begin{figure}[t]
    \centering
    \includegraphics[width=\textwidth]{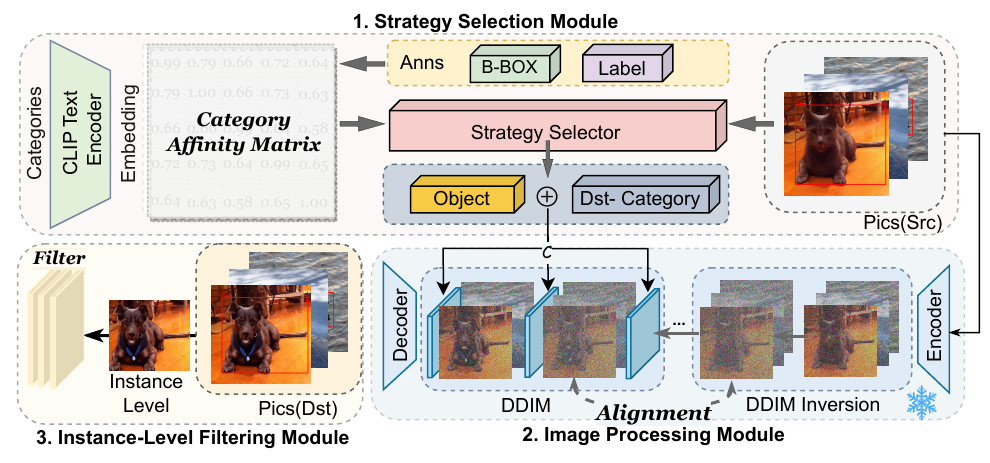}
    \caption{Overview of our method. 
    Step 1 constructs the Category Affinity Matrix to develop tailored augmentation strategies for enhancing data diversity in each image.
    In Step 2, 
    the method generates diverse images under the guidance of the Strategy Selection Module while maintaining semantic coordination through Surrounding Region Alignment with a diffusion model.
    Step 3 involves the exclusion of low-quality images at the instance level to further ensure the overall dataset quality.
    }
    \label{fig:overviewnew}
\end{figure}


\section{Method}
In this paper, 
we endeavor to devise a data augmentation method that harmonizes image diversity and semantic coordination for object detection. 
Given an original dataset $ \mathcal{D} = { (I_1, \mathcal{B}_1, \mathcal{L}_1),...,(I_N, \mathcal{B}_N, \mathcal{L}_N) } $ of $N$ images, each $I_i$ with manually labeled bounding boxes $\mathcal{B}_i$ and corresponding labels $\mathcal{L}_i$, 
our method is to steer a diffusion model into a dataset synthesizer and produce an augmented dataset $\mathcal{D'}$.
As depicted in Fig.~\ref{fig:overviewnew},
the overall framework consists of three major modules: 
$\textbf{\textit{(1)}}$ Strategy Selection Module.
Initiating with the development of a Category Affinity Matrix, this module quantifies inter-class relationships, facilitating the informed selection of an optimal object and a target category for each image. 
This targeted approach directs the diffusion model to generate images that are diverse yet contextually coordinated.
$\textbf{\textit{(2)}}$ Image Processing Module.
This module incorporates a Surrounding Region Alignment strategy, ensuring that the augmentation process preserves the semantic context of the entire image, thus maintaining overall image semantic coordination.
$\textbf{\textit{(3)}}$ Instance-Level Filtering Module. 
To further uphold the augmented dataset’s quality, 
We employ instance-level filtering to filter out images of low quality.
In the following sections, we will elaborate on each module in detail.

\subsection{Strategy Selection Module} \label{Strategy Selection Module}
\subsubsection{Category Affinity Matrix.}
To calculate the category affinity of two categories,
we extract their visual $v$ and semantic $l$ features.
This extraction process usually requires training on large image datasets. 
Fortunately, the CLIP model ~\cite{radford2021learning} is trained on large text-image pairs and has proven capabilities in image feature extraction and semantic understanding. 
The training goal of CLIP, 
to correlate text and image features, 
allows us to view the semantic feature embeddings $E(l)$ as proxies for their corresponding visual features $E(v)$.
Consequently, $E(l)$ embodies a comprehensive mix of rich visual and semantic information.
We then construct the category affinity matrix $A$ by computing the cosine similarity between the embeddings of each category pair.
In this matrix, 
$A_{i,j}$ represents the affinity between categories $l_i$ and $l_j$, and it can be expressed as:
\begin{equation}
A_{ij} = \frac{\mathbf{E}(l_{i}) \cdot \mathbf{E}(l_{j})} {\|\mathbf{E }(l_{i})\| \|\mathbf{E}(l_{j})\|} \approx \frac{\mathbf{E}(v_{i}) \cdot \mathbf{E}(v_ {j})} {\|\mathbf{E}(v_{i})\| \|\mathbf{E}(v_{j})\|}.
\end{equation}

\subsubsection{Optimal Object Identification.}
Images often contain multiple objects, so it is necessary to select an appropriate object $o_k$ for editing based on various factors. 
In our method, we take object category and size into account and determine selection probabilities using category and area scores. 
The category score $C_{1}(o_k)$ measures the category affinity of the object with other objects, calculated as the sum of the affinities with all other categories. 
The area score $C_{2}(o_k)$ corresponds to the size of the object. 
Objects that occupy only a small part of the image may limit the editable area, while those that dominate the image may excessively modify the background, both undesirable outcomes. To strike a balance, we define $ \alpha $ as the best selection ratio.
Thus, $C_{2}(o_k)$ can be calculated by the object's size $ s(o_k) $ and the image size $ s(I)$:
\begin{equation}
    C_{1}(o_k) =  {\textstyle \sum_{j=1}^{n}} A_{k,j}, \quad C_{2}(o_k) = 1 - \frac{\left |  s(o_k) -s(I) \cdot \alpha \right |}{ s(I) }.
\end{equation}
The probability $ p(o_k) $ depends on Min-Max normalized $C_{1}(o_k)$ and $C_{2}(o_k)$ :

\begin{equation}
p(o_k)= \frac{e^{\left |   \hat{C}_{1}(o_k)  + \hat{C}_{2}(o_k) \right |}}{ {\sum_{i=1}^{n}} e^{\left | \hat{C}_{1}(o_i)  + \hat{C}_{2}(o_i)\right |}}. 
\end{equation}

We then sample from the probability distribution to select the target object.




\subsubsection{Target Category Determination.}
Selecting a target category $\mathcal{C}_{dst}$ involves identifying a category into which the object $o_k$ will be transformed. 
Utilizing the Category Affinity Matrix, which enumerates potential categories, 
the matrix element $A_{i,j}$ quantifies the likelihood of targeting category $l_j$.
When $\mathcal{C}_{dst}$ aligns with the original category $\mathcal{C}_{src}$,
our approach regenerates the object to enhance inter-class variability.
Conversely, if $\mathcal{C}_{dst}$ differs, 
it facilitates the transformation of the object into the designated category, thereby augmenting inter-class diversity. 
To optimize the selection process, we integrate two additional criteria:
$\textbf{\textit{(1)}}$ A threshold $\theta$ is established to exclude categories with affinity below this threshold from being targeted, mitigating the potential for semantic dissonance.  
$\textbf{\textit{(2)}}$ The probability of selecting the original category 
$\mathcal{C}_{src}$ as the target is diminished, which strategically increases the diversity of the augmented dataset.


\subsection{Image Processing Module} \label{Image Processing Module}

\begin{figure}[ht]
    \centering
    \includegraphics[width=\textwidth]{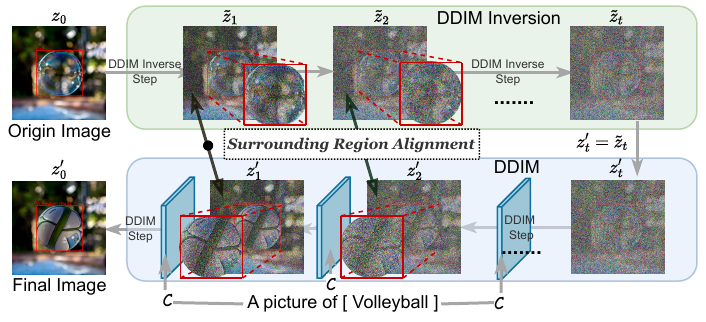}
    \caption{The process of the Image Processing Module. $\textbf{\textit{(1)}}$ We get the initial noise $\tilde{z}_t $ through DDIM inversion. 
    $\textbf{\textit{(2)}}$ We take the Surrounding Region Alignment in the environment region and conditional control editing in the object region.}
    \label{fig:overview}
\end{figure}

\vspace{-\baselineskip}

\subsubsection{DDIM Inversion.}



Image generation typically begins by randomly sampling noise $ \mathbf{z}_T$ from a Gaussian distribution during the diffusion process. 
However, to preserve the original information of the image, we employ a methodology akin to image editing. 
Instead of starting from random noise, we start from the initial noise $\tilde  \mathbf{z}_T$ using specialized inversion steps called DDIM inversion, which contains information from the original image.
Fellow previous researches~\cite{huang2024diffusion,song2020denoising}, we can obtain the form of DDIM Inversion as follows:
\begin{equation}
    \tilde {\mathbf{z}}_{t}=\sqrt{\bar{\alpha}_{t}} \frac{\tilde \mathbf{z}_{t-1}-\sqrt{1-\bar{\alpha}_{t-1}} \epsilon_{\theta}\left(\tilde \mathbf{z}_{t-1}, t\right)}{\sqrt{\bar{\alpha}_{t-1}}}+\sqrt{1-\bar{\alpha}_{t}} \epsilon_{\theta}\left(\tilde \mathbf{z}_{t-1}, t\right),
\end{equation}
where $\epsilon_{\theta}$ represents the diffusion process parameterized by $\theta$, $ \tilde {\mathbf{z}}_{t} $ signifies the noise at time $t$, and $\bar{\alpha}$ is the hyperparameter of the diffusion process. In particular, our DDIM inversion process information is captured and used for subsequent Surrounding Region Alignment work.

\subsubsection{Text-conditional Image Editing.}
In the previous section, we got the target category $\mathcal{C}_{dst}$, which is structured as a prompt: "A picture of [$ \mathcal{C}_{dst} $]". This prompt serves as the control condition $C$ for the diffusion model. Based on the previous research, the Stable Diffusion\cite{rombach2022high} model has been thoroughly trained on extensive datasets, so our experiments build on this pre-trained model.
The entire process of conditional control can be articulated as follows:
\begin{equation}
    \epsilon_{\theta}\left(\mathbf{z'}_{t}, t, C, \varnothing\right)=w \cdot \epsilon_{\theta}\left(\mathbf{z'}_{t}, t, C\right)+(1-w) \cdot \epsilon_{\theta}\left(\mathbf{z'}_{t}, t, \varnothing\right),
\end{equation}
where $C$ denotes the control condition, $ \mathbf{z'}_{t} $ signifies the new noise at time $t$, and $\varnothing$ indicates no condition. The parameter $w$ modulates the contribution of the control condition $C$ relative to the absence of any condition.

\subsubsection{Surrounding Region Alignment.}


During the DDIM inversion process, we obtain the image information $\tilde \mathbf{z}_t$ at different time steps. 
This set of information is employed to modify the noise $\mathbf{z'}_{t}$ in the denoising process. In the next denoising process, the diffusion model coordinates the original environment region and the new object region to make the whole image semantically coordinated.
At each time step, the Surrounding Region Alignment is formulated as:
\begin{equation}
    \mathbf{z'}_{t} \leftarrow \mathbf{z'}_{t} *Area_{o_k} + \tilde \mathbf{z}_{t} *(1-Area_{o_k}).
\end{equation}
Here, $Area_{o_k}$ represents the mask corresponding to the object region.


\subsection{Instance-Level Filtering Module}
\label{Instance-Level Filtering Module}
To obtain higher-quality generated data, we need to filter the images generated. 
In our endeavor, we confine our attention solely to the object region while leaving the environment region unchanged, thus necessitating a filtering approach at the instance level. 
Our method requires segmenting out the object region and classifying it with a pre-trained classifier~\cite{he2016deep}.
Subsequently, we extract the top $k$ predictions yielded by the classifier. 
If these predictions are all different from the target category $\mathcal{C}_{dst}$, the generated data is considered of poor quality and therefore filtered out.


\section{Experiment}


\subsection{Experimental Settings}

\noindent \textbf{Datasets.} In our research, we utilize three widely recognized datasets to conduct our analysis: Microsoft COCO~\cite{lin2014microsoft} of approximately 330,000 images, Objects365~\cite{shao2019objects365} with around 2,000,000 images (we only focused on food categories), and Open Images~\cite{OpenImages} of approximately 19,000,000 images. These datasets serve as comprehensive samples, representing standard, category-specific, and fine-grained data respectively.


\noindent \textbf{Detectors.} We conduct experiments on three detectors, including Faster R-CNN~\cite{ren2016faster}, Mask R-CNN~\cite{he2018mask}, and YOLOX~\cite{ge2021yolox}. They include both single-stage and two-stage detection methods, both of which are widely used in object detection. In subsequent experiments, the Resnet-50 is used as the backbone.

\begin{table}[t]
    \centering
    \caption{Results on 50\% COCO dataset with six data augmentation methods. $R$ denotes categories that have no affinity with others,
    i.e. $R$ have the same $\mathcal{C}_{dst}$ and $\mathcal{C}_{src}$. $T$ denotes categories that share an affinity with others.}
    
    \label{tab:main-res-1}
    \renewcommand\arraystretch{0.8}
    \begin{tabular}{lclll}
        \toprule
        \hspace{0.5cm}Detector & \multicolumn{1}{c}{\hspace{0.4cm}Method} & \multicolumn{1}{c}{\hspace{0.7cm}$AP_{50}^{R}$} & \multicolumn{1}{c}{\hspace{0.7cm}$AP_{50}^{T}$} & \multicolumn{1}{c}{\hspace{0.7cm}$AP_{50}^{All}$} \\ 
        \cmidrule(lr){1-2}\cmidrule(lr){3-5}
        \multirow{6}{*}{\hspace{0.1cm}Faster R-CNN~\cite{ren2016faster}} &\hspace{0.5cm}Vanilla        &\hspace{1cm}{32.4}  &\hspace{1cm}{31.9}  &\hspace{1cm}32.2                         \\ 
        & \hspace{0.5cm}Random Erasing        &\hspace{1cm}{33.5}  &\hspace{1cm}{32.0}  &\hspace{1cm}33.2                         \\
                                      & \hspace{0.5cm}MixUp                 &\hspace{1cm}{33.6}  &\hspace{1cm}{33.9}  &\hspace{1cm}33.8                         \\
                                      & \hspace{0.5cm}Mosaic                &\hspace{1cm}{35.0}  &\hspace{1cm}{33.0}  &\hspace{1cm}34.3                         \\
        \cdashline{2-5}                              
                                      & \hspace{0.5cm}SD-Inpaint            &\hspace{1cm}{34.8}  &\hspace{1cm}{34.0}  &\hspace{1cm}34.6                         \\
        & \hspace{0.5cm}X-Paste            &\hspace{1cm}{34.7}  &\hspace{1cm}{35.1}  &\hspace{1cm}34.9                         \\
                                      & \hspace{0.5cm}\textbf{Ours}         &\hspace{0.95cm}\textbf{{35.7}}  &\hspace{0.95cm}\textbf{{35.3}} &\hspace{0.95cm}\textbf{35.6}                        \\ 
        \cmidrule(lr){1-5}
        \multirow{6}{*}{\hspace{0.2cm}Mask R-CNN~\cite{he2018mask}}   &\hspace{0.5cm}Vanilla        &\hspace{1cm}{33.6}  &\hspace{1cm}{34.2}  &\hspace{1cm}33.8                         \\ 
        & \hspace{0.5cm}Random Erasing        &\hspace{1cm}{33.8}     &\hspace{1cm}{34.4}               &\hspace{1cm}34.0                         \\
                                      & \hspace{0.5cm}MixUp                &\hspace{1cm}{35.0}     &\hspace{0.95cm}{\textbf{35.9}}               &\hspace{1cm}35.3                         \\
                                      & \hspace{0.5cm}Mosaic                &\hspace{1cm}{35.3}     &\hspace{1cm}{35.0}               &\hspace{1cm}35.2                         \\
        \cdashline{2-5}                              
                                      & \hspace{0.5cm}SD-Inpaint            &\hspace{0.95cm}{\textbf{36.5}}   &\hspace{1cm}{35.3}               &\hspace{1cm}\textbf{36.1}                         \\
                                      & \hspace{0.5cm}X-Paste           &\hspace{1cm}{35.6}    &\hspace{1cm}{35.1}               &\hspace{1cm}35.4                         \\
                                      & \hspace{0.5cm}\textbf{Ours}         &\hspace{1.0cm}{36.4}     &\hspace{1.0cm}{35.4}               &\hspace{1.0cm}\textbf{36.1}                       \\ 
        \cmidrule(lr){1-5}
        \multirow{6}{*}{\hspace{0.55cm}YOLOX~\cite{ge2021yolox}}        &\hspace{0.5cm}Vanilla        &\hspace{1cm}{30.5}  &\hspace{1cm}{31.1}  &\hspace{1cm}30.7                         \\ 
        & \hspace{0.5cm}Random Erasing        &\hspace{1cm}{31.0}     &\hspace{1cm}{32.4}               &\hspace{1cm}31.5                         \\
                                      & \hspace{0.5cm}MixUp                 &\hspace{1.25cm}\--{}     &\hspace{1.25cm}\--{}               &\hspace{1.25cm}\--{}                        \\
                                      & \hspace{0.5cm}Mosaic                &\hspace{1.25cm}\--{}     &\hspace{1.25cm}\--{}               &\hspace{1.25cm}\--{}                         \\
        \cdashline{2-5}                              
                                      & \hspace{0.5cm}SD-Inpaint            &\hspace{1cm}{35.5}     &\hspace{1cm}{36.6}               &\hspace{1cm}35.9                         \\
                                      & \hspace{0.5cm}X-Paste            &\hspace{1cm}{33.5}     &\hspace{1cm}{34.0}               &\hspace{1cm}33.7                         \\
                                      & \hspace{0.5cm}\textbf{Ours}         &\hspace{0.95cm}{\textbf{36.8}}     &\hspace{0.95cm}{\textbf{37.9}}               &\hspace{0.95cm}\textbf{37.1}                         \\ 
        \bottomrule
    \end{tabular}
\end{table}

\noindent \textbf{Baselines.} We compared three typical model-free methods: Random Erase~\cite{zhong2020random}, Mixup~\cite{zhang2017mixup}, and Mosaic~\cite{ge2021yolox}, and two novel model-based methods: SD-inpaint~\cite{fang2024data} and X-Paste~\cite{zhao2023x} with our method. 

\noindent \textbf{Implementation Details.} We set the affinity threshold $\theta$ for COCO, Objects365, and Open Images at the top of the affinity scale: 3\%, 2.5\%, and 0.25\%, with the best selection ratio $\alpha$ at 0.35. For the matrix, we use the pre-trained CLIP model~\cite{radford2021learning}. For the diffusion, we employ the pre-trained stable diffusion model with $w$ value of 7.5, following the original paper \cite{rombach2022high}. For the filter, we choose the ResNet-50~\cite{he2016deep} as an instance-level filter with $k$ set to 3.

\subsection{Main Results}\label{Main Results}
In this section, we conduct experiments on three detectors with six data augmentation methods on three datasets. Our main results are presented in Table \ref{tab:main-res-1} and Table \ref{tab:main-res-2}. We analyze them from two perspectives: $\textbf{\textit{(1)}}$ detector-specific performance; $\textbf{\textit{(2)}}$ dataset-specific performance.


\noindent \textbf{Detector-specific Performance.} 
Our method shows notable performance improvements on all benchmark detectors, covering both single-stage and two-stage architectures. 
As presented in Table~\ref{tab:main-res-1}, we achieved improvements of +\textbf{3.4}AP, +\textbf{2.3}AP, and +\textbf{6.4}AP on the Faster R-CNN~\cite{ren2016faster}, Mask R-CNN~\cite{he2018mask}, and YOLOX~\cite{ge2021yolox} models compared to non-augmentation (referred to as "vanilla").
This demonstrates that our method has a beneficial effect on different detector architectures.
Notably, our method significantly improves YOLOX, which already incorporates other methods, demonstrating our potential to further improve performance in combination with existing methods.
Comparatively, our method achieved an average advantage of +\textbf{1.4}AP, +\textbf{0.9}AP, and +\textbf{3.4}AP over alternative approaches. This highlights the effectiveness of our strategy in balancing diversity and semantic coordination, thereby facilitating improved learning from data-driven models.

\vspace{-\baselineskip}
 \begin{table}[htbp]
    \centering
    \caption{Results on subsets of Objects365 Food and Open Images with six data augmentation methods. We use the AP50 metric on Faster R-CNN to evaluate performance. The description of $R$ and $T$ is the same as Table \ref{tab:main-res-1}.}
    \label{tab:main-res-2}
    \renewcommand\arraystretch{0.9}
    \begin{tabular}{lclll}
        \toprule
        \hspace{1cm}Dataset & \multicolumn{1}{c}{\hspace{0.4cm}Method} & \multicolumn{1}{c}{\hspace{0.7cm}$AP_{50}^{R}$} & \multicolumn{1}{c}{\hspace{0.7cm}$AP_{50}^{T}$} & \multicolumn{1}{c}{\hspace{0.7cm}$AP_{50}^{All}$} \\ 
        \cmidrule(lr){1-2}\cmidrule(lr){3-5}
        \multirow{6}{*}{\hspace{0.15cm} Objects365 Food~\cite{shao2019objects365}} &\hspace{0.5cm}Vanilla        &\hspace{1cm}{20.8}  &\hspace{1cm}{17.3}  &\hspace{1cm}19.1                         \\ 
        & \hspace{0.5cm}Random Erasing        &\hspace{1cm}{22.0}  &\hspace{1cm}{19.6}  &\hspace{1cm}20.9                         \\
                                      & \hspace{0.5cm}MixUp                 &\hspace{1cm}{21.6}  &\hspace{1cm}{18.6}  &\hspace{1cm}20.2                         \\
                                      & \hspace{0.5cm}Mosaic                &\hspace{1cm}{22.4}  &\hspace{1cm}{20.2}  &\hspace{1cm}21.4                         \\
        \cdashline{2-5}                              
                                      & \hspace{0.5cm}SD-Inpaint            &\hspace{1cm}{22.2}  &\hspace{1cm}{20.5}  &\hspace{1cm}21.4                         \\
        & \hspace{0.5cm}X-Paste            &\hspace{1cm}{21.1}  &\hspace{1cm}{19.1}  &\hspace{1cm}20.2                         \\
                                      & \hspace{0.5cm}\textbf{Ours}         &\hspace{0.95cm}\textbf{{23.2}}  &\hspace{0.95cm}\textbf{{22.4}} &\hspace{0.95cm}\textbf{22.8}                        \\ 
        \cmidrule(lr){1-5}
        \multirow{6}{*}{\hspace{0.05cm} Open Images~\cite{OpenImages}}   &\hspace{0.5cm}Vanilla        &\hspace{1cm}{19.1}  &\hspace{1cm}{14.0}  &\hspace{1cm}17.6                         \\ 
        & \hspace{0.5cm}Random Erasing        &\hspace{1cm}{21.0}     &\hspace{1cm}{16.8}               &\hspace{1cm}19.8                         \\
                                      & \hspace{0.5cm}MixUp                &\hspace{1cm}{20.7}     &\hspace{1cm}{15.6}               &\hspace{1cm}19.2                         \\
                                      & \hspace{0.5cm}Mosaic                &\hspace{1cm}{22.2}     &\hspace{1cm}{19.6}               &\hspace{1cm}21.4                         \\
        \cdashline{2-5}                              
                                      & \hspace{0.5cm}SD-Inpaint            &\hspace{1cm}{22.5}   &\hspace{0.95cm}{\textbf{19.9}}               &\hspace{1cm}{21.7}                         \\
                                      & \hspace{0.5cm}X-Paste           &\hspace{1cm}{21.5}    &\hspace{1cm}{16.2}               &\hspace{1cm}19.9                         \\
                                      & \hspace{0.5cm}\textbf{Ours}         &\hspace{0.95cm}{\textbf{23.2}}     &\hspace{1.0cm}{19.0}               &\hspace{0.95cm}\textbf{22.0}                       \\ 
        
        \bottomrule
    \end{tabular}
\end{table}
\vspace{-\baselineskip}


\noindent \textbf{Dataset-specific Performance.} 
Our method demonstrates robust applicability to three types of datasets (standard, category-specific, and fine-grained), achieving optimal or comparable performance. 
For each dataset, a subset of 30\% of it was selected for the efficiency of the test. 
The results are detailed in Table \ref{tab:main-res-1} and Table \ref{tab:main-res-2}.
Compared to vanilla, our approach achieves a performance improvement of +\textbf{3.4}AP, +\textbf{3.6}AP, and +\textbf{4.4}AP on three types of datasets. 
Particularly, on category-specific and fine-category datasets, our method outperforms other methods by +\textbf{1.98}AP and +\textbf{1.6}AP on average. 
This is because the Category Affinity Matric can select more affinity categories as target categories in category-specific and fine-category datasets. 
Consequently, it achieves a more balanced amalgamation of diversity and semantic coordination, thereby contributing to overall model improvement. 
For an in-depth analysis, please refer to Section \ref{Diversity and Semantic Coordination}.

\subsection{Diversity and Semantic Coordination} \label{Diversity and Semantic Coordination}

In this section, we aim to uncover the underlying factors contributing to the superior performance of our method. Specifically, we will discuss diversity and semantic coordination from three perspectives \cite{li2023semantic}: $\textbf{\textit{(1)}}$ human evaluation; $\textbf{\textit{(2)}}$ similarity evaluation; $\textbf{\textit{(3)}}$ case study.

\begin{table}[t]
\centering
\caption{Human evaluation results of our data augmentation method and others on three datasets from the perspectives of image diversity (Div.) and semantic coordination (Coo.)}.
\label{tab:human}
\renewcommand\arraystretch{0.8}
\begin{tabular}{@{}l@{\hspace{10pt}}*{8}{>{\centering\arraybackslash}p{1.1cm}}@{}}
\toprule
\multirow{2}{*}{Method} & \multicolumn{2}{c}{COCO} & \multicolumn{2}{c}{Objects365} & \multicolumn{2}{c}{Open images} & \multicolumn{2}{c}{Avg.} \\ \cmidrule(lr){2-3} \cmidrule(lr){4-5} \cmidrule(lr){6-7} \cmidrule(lr){8-9}
                         & Div.       & Coo.      & Div.           & Coo.           & Div.            & Coo.            & Div.    & Coo.                  \\ \midrule
Random Erasing             & 1.25       & 3.77      & 1.29           & 3.36           & 1.22            & 3.96            & 1.25           & 3.70                  \\
MixUp                   & \textbf{3.44}       & 1.68      & 3.33           & 1.53           & 3.3             & 1.64            & 3.36           & 1.62                  \\
Mosaic                   & 2.76       & 2.23      & 2.97           & 2.50            & 2.82            & 2.02            & 2.85           & 2.25                  \\
SD-Inpaint               & 1.33       & \textbf{4.59}      & 1.22           & \textbf{4.33}           & 1.14            & \textbf{4.61}            & 1.23          & \textbf{4.51}                  \\
X-Paste                  & 4.40        & 1.29      & \textbf{4.28}           & 1.17           & \textbf{4.51}            & 1.36            & \textbf{4.40}           &  1.27                 \\
Ours                     & 3.12       & 2.96      & 2.71           & 2.61           & 2.77            & 3.26            & 2.87           &  2.94                 \\ \bottomrule
\end{tabular}
\end{table}

\begin{figure}[t]
    \centering
    \begin{minipage}[b]{0.60\linewidth}
        \centering
        \includegraphics[width=\textwidth]{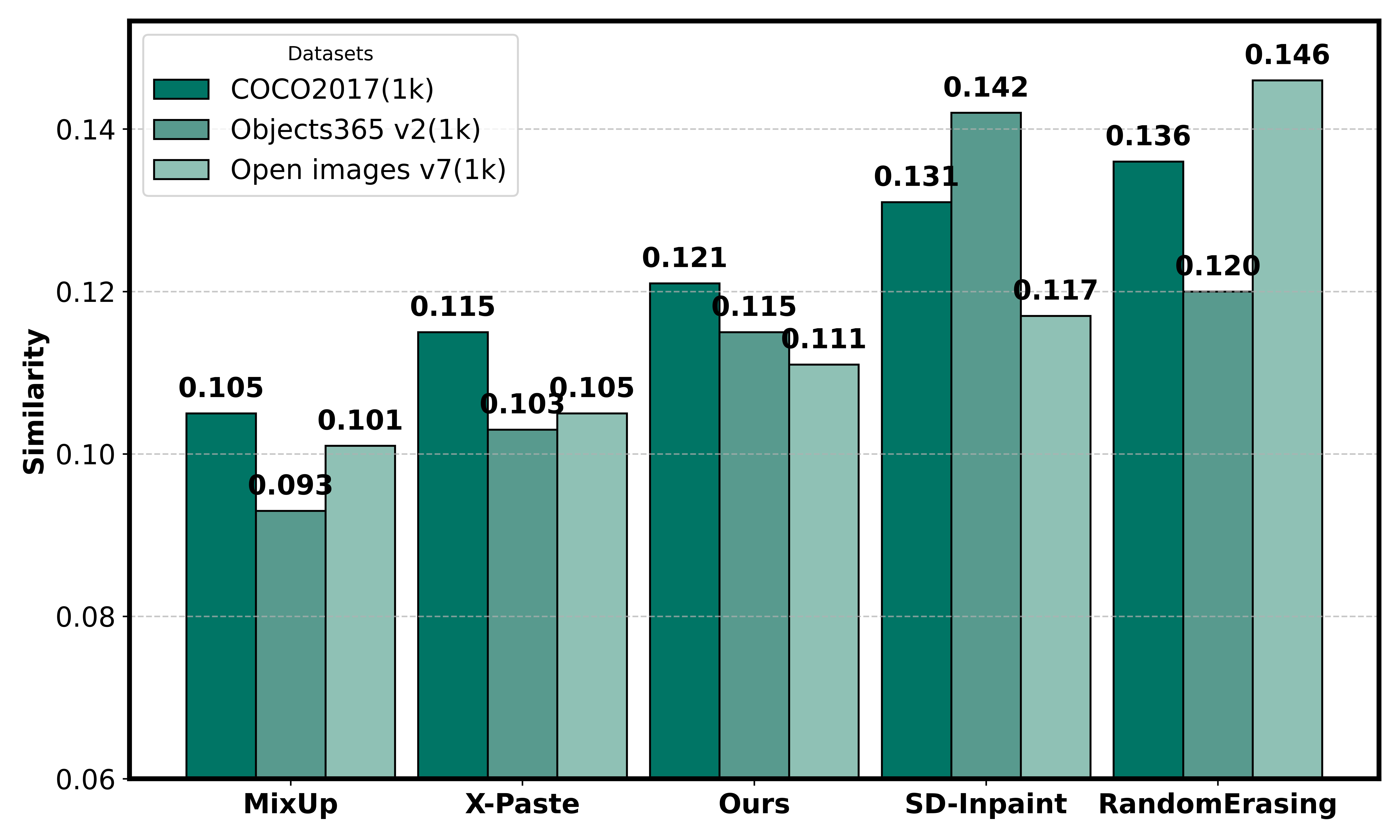}
        \caption{Average cosine similarity between the augmented and original images on subsets of COCO, Objects365, and Open images.}
        \label{fig:similarity}
    \end{minipage}
    \hfill
    \begin{minipage}[b]{0.36\linewidth}
        \centering
        \includegraphics[width=\textwidth]{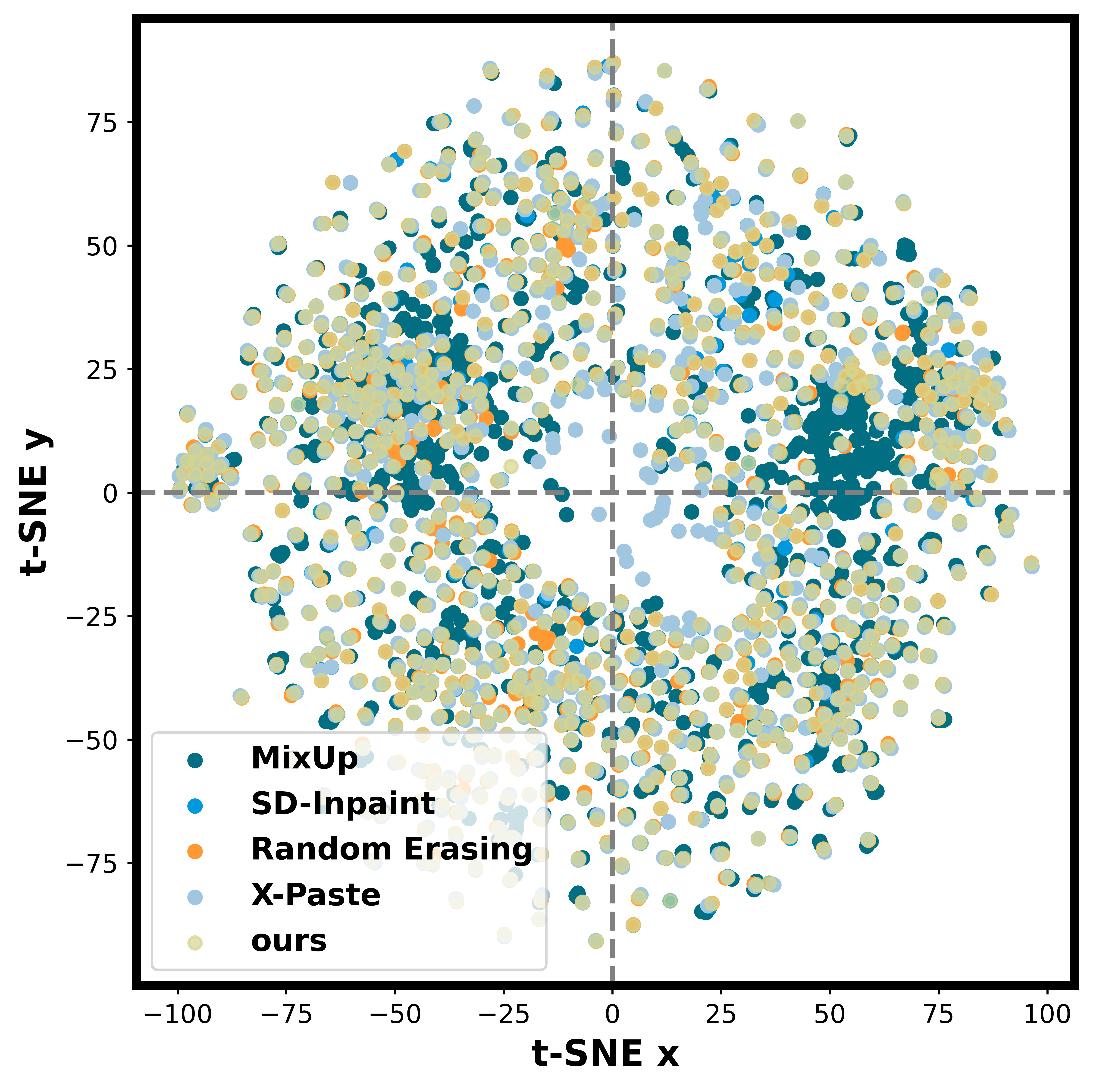}
        \caption{Distribution of augmented datasets with our method and others.}
        \label{fig:t-SNE}
    \end{minipage}
\end{figure}

\noindent \textbf{Human Evaluation.} 
We apply human evaluation on COCO, Object365, and Open Images. For each dataset, we randomly choose 10 labels and 10 of their corresponding original images. We evaluate the augmented images by six data augmentation methods based on the original image. Each augmented image is scored on a scale of 1 to 5 in terms of image diversity and semantic consistency respectively \cite{li2023semantic}. We employ eight annotators who are trained and pass trial annotations. Table~\ref{tab:human} shows the human evaluation results for six methods and their corresponding average performance.

Based on the results, our method demonstrates comparable semantic coordination and greater diversity than Random Erase and SD-inpaint, while exhibiting higher semantic coordination and lower diversity than Mix-up and X-paste. When considering object detection performance, our method effectively balances image diversity and semantic coordination, leading to the highest performance. 

\begin{figure}[t]
    \centering
    \includegraphics[width=0.95\textwidth]{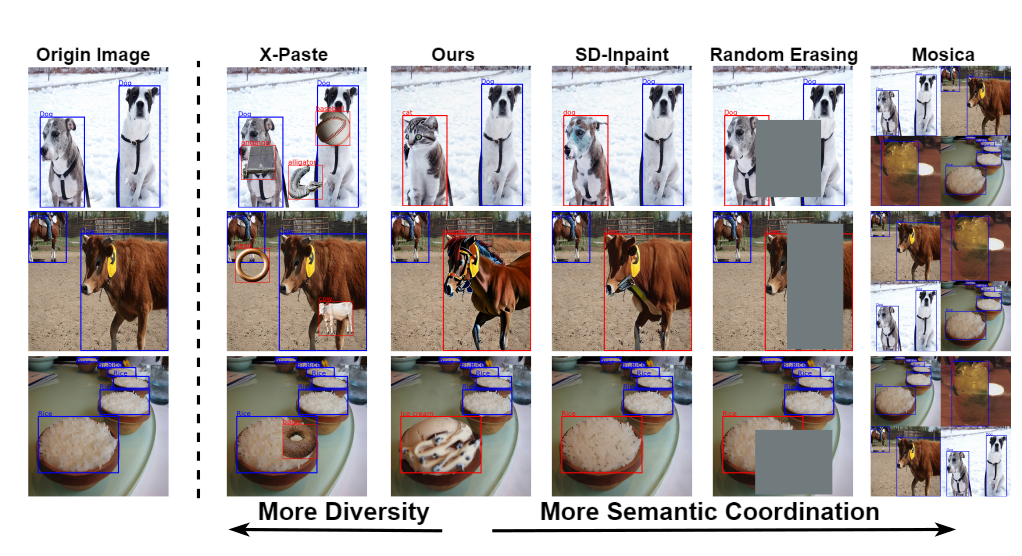}
    \caption{Case study of different data augmentation methods.} 
    \label{fig:case_study}
\end{figure}

\noindent \textbf{Similarity Evaluation.} 
We conducted a similarity evaluation on subsets of three datasets, each comprising 1,000 images,  employing six distinct data augmentation methods. For each method, we computed the cosine similarity between the original and augmented images, subsequently averaging the scores across the entire dataset to quantify diversity. A lower average similarity implies increased diversity and a higher potential for semantic inconsistency.~\cite{li2023semantic} As depicted in Fig. \ref{fig:similarity}, our findings reveal that our method exhibits intermediate results, moderate cosine similarity (0.121), outperforming others in maintaining this balance. The distribution displayed in Fig.\ref{fig:t-SNE} further supports this observation, with our method showing moderate dispersion, emphasizing the significance of striking a harmony between image diversity and semantic coordination.


\noindent \textbf{Case Study.} 
Fig. \ref{fig:case_study} shows the images generated by our method and others. Three of these methods introduce slight modifications to augment the original images. 
They redraw, mask, or splice images, but none of them sufficiently diversify the original images. 
Although X-paste introduces random objects, producing diverse and appealing results, it often disrupts semantic coordination.
For example, the introduction of a floating alligator between two dogs.
In contrast, our method preserves the semantic integrity of the whole original image and achieves diversity in the augmented images. 
For example, transformations like a dog becoming a cat or a cow turning into a horse demonstrate our method's ability to introduce meaningful variations.







\subsection{Ablation Study}

In this section, to understand the effectiveness of the proposed components, we perform ablation studies on the subset of the Objects365 food category, investigating the effect of the Category Affinity Matrix, Surrounding Region Alignment, and the Instance-Level Filter.



\begin{figure}[t]
    \centering
    \begin{minipage}[t]{0.50\linewidth}
        \vspace{5pt}
        \centering
        \renewcommand\arraystretch{0.9}
        \begin{tabular}{*{4}{>{\centering\arraybackslash}p{1.4cm}}}
            \toprule
            Matrix & Alig. & Filter & $AP_{50}^{All}$ \\
            \cmidrule(lr){1-3} \cmidrule(lr){4-4}
             \XSolidBrush & \XSolidBrush & \XSolidBrush & 5.2 \\
            \Checkmark & \XSolidBrush & \XSolidBrush & 6.0 \\
            \XSolidBrush & \Checkmark &  \XSolidBrush & 7.7 \\
            \Checkmark & \Checkmark & \XSolidBrush & 11.7 \\
            \Checkmark & \Checkmark & \Checkmark & 12.2 \\
            \bottomrule
        \end{tabular}
        \captionof{table}{The effectiveness of our proposed Matrix, strategy, and instance-level filter. Matrix and Alig. refer to Category Affinity Matrix and Surrounding Region Alignment.}
        \label{tab:components}
    \end{minipage}
    \hfill
    \begin{minipage}[t]{0.42\linewidth}
        \vspace{0pt}
        \centering
        \includegraphics[width=\linewidth]{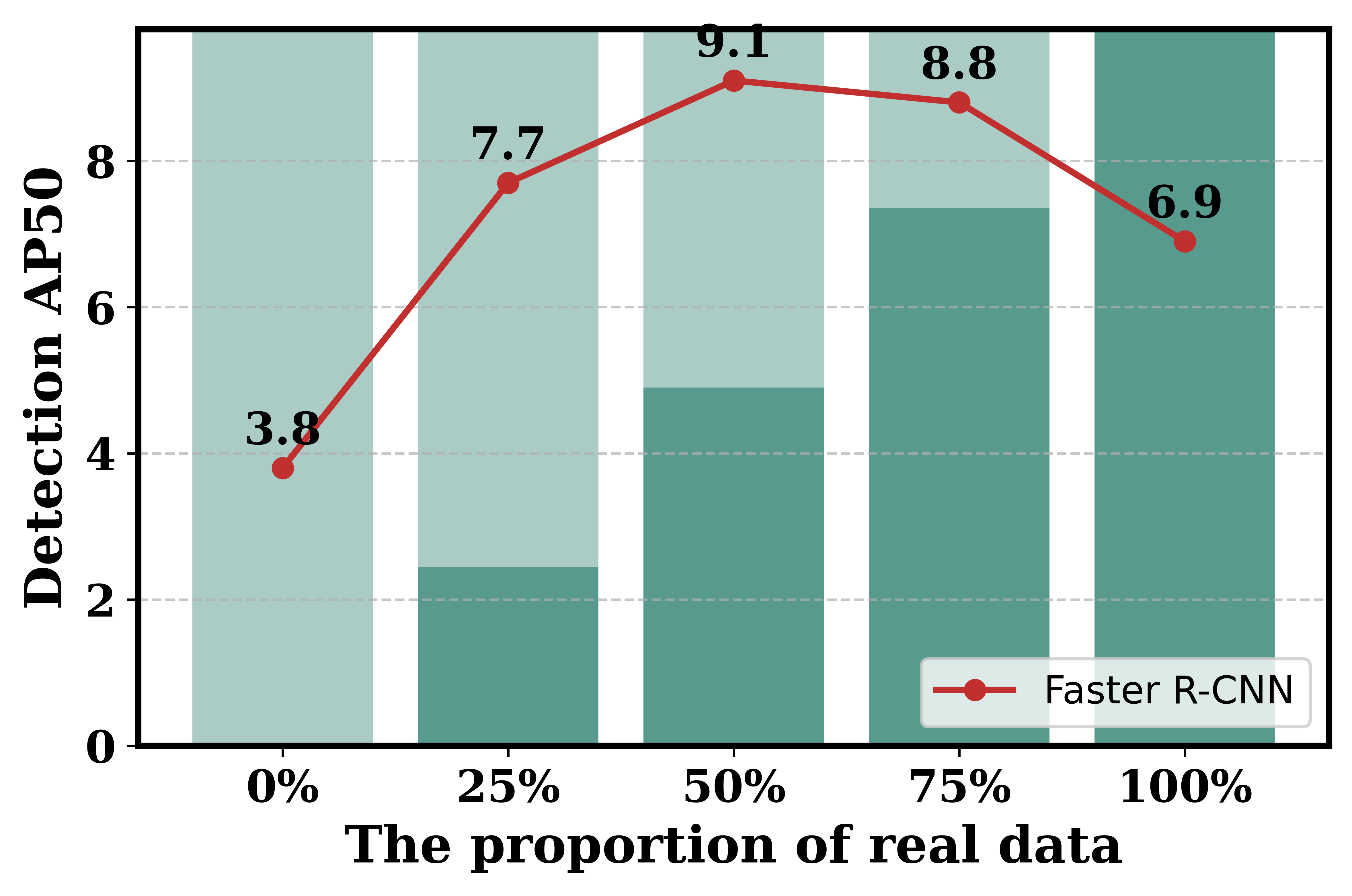}
        \caption{AP50 with rising real data proportions. Dark green represents real data and Light green represents augmented data.}
        \label{fig:proportion}
    \end{minipage}
\end{figure}

\noindent \textbf{Category Affinity Matrix.}
We assess the effectiveness of the Category Affinity Matrix and its impact on generative images for training object detectors. We employed the generated images to train Faster R-CNN for object detection. The results are presented in Table~\ref{tab:components}, showing that in the same case with and without Surrounding Region Alignment, the Category Affinity Matrix delivers improvement detection performance by +4AP and +0.8AP. This indicates that the Matrix plays an important role in affecting the performance of object detection, and further reflects its effectiveness in augmenting data diversity.

\noindent \textbf{Surrounding Region Alignment.}
To demonstrate the effectiveness of the Surrounding Region Alignment, we evaluate the detection performance with generated images. As shown in Table~\ref{tab:components}, training Faster R-CNN with generated images, leads to a noticeable enhancement in detection performance, increasing from 6.0 and 5.2 to 11.7(+5.7AP) and 7.7(+2.5AP), respectively. In the same way, this indicates that the Surrounding Region Alignment also plays an important role in affecting the performance of object detection, and further reflects its effectiveness in coordinating the semantics of the data.

\noindent \textbf{Instance-Level Filter.}
We further evaluate the detection performance after introducing the Instance-Level Filter. As shown in Table~\ref{tab:components}, when introducing the Instance-Level Filter while reserving Category Affinity Matrix and Surrounding Region Alignment, the detector's performance increases from 11.7 to 12.2(+0.5AP), showing that the Instance-Level Filter also delivers improvement on detection performance. 



\section{Discussion}


\noindent \textbf{Influence of Real Data Percentage.}
We investigated the influence of varying proportions of real data on the performance of the Faster-RCNN using COCO datasets(5k). The total data volume for the experiment was kept constant, and the results are illustrated in Fig. \ref{fig:proportion}. Generally, increasing the proportion of real data significantly enhances the model's performance. However, beyond a certain threshold, further increases in real data proportion lead to a decline in model performance. This indicates that while more real data is beneficial for model training, it is also necessary to incorporate some generated data to optimize performance effectively.

\noindent \textbf{Influence of Affinity Threshold.}
In our methodology, the affinity threshold $\theta$ affects the quality of the dataset. Lower affinity thresholds increase the likelihood that objects will be transformed into other categories, while higher affinity thresholds cause the dataset to maintain its original semantics more. In particular, when the affinity threshold is higher than all affinities, our method will be similar to redrawing local regions of the image. In our experiments, we determined the threshold by manually tuning the parameters. In future work, how to adaptively adjust the affinity threshold according to different datasets is a question worth continuing to investigate.

\section{Conclusion}

In this paper, we propose a novel diffusion model-based method for data augmentation in object detection, aiming to boost image diversity while maintaining semantic coordination. Our method incorporates three key components: a Category Affinity Matrix to optimize diversity, a Surrounding Region Alignment strategy for semantic coordination, and an Instance-Level Filter to further enhance dataset quality. Experimental findings demonstrate that our method strikes a balance between semantic coordination and image diversity, resulting in promising outcomes for object detection.

\makeatletter
\newcommand\figcaption{\def\@captype{figure}\caption}
\newcommand\tabcaption{\def\@captype{table}\caption}
\makeatother

\bibliographystyle{splncs04}
\bibliography{mybibliography}
%





\end{sloppypar}

\end{document}